\documentclass[sigconf]{acmart}  
\AtBeginDocument{%
  \providecommand\BibTeX{{%
    \normalfont B\kern-0.5em{\scshape i\kern-0.25em b}\kern-0.8em\TeX}}}

\copyrightyear{2023}
\acmYear{2023}
\setcopyright{acmlicensed}\acmConference[MM '23]{Proceedings of the 31st ACM International Conference on Multimedia}{October 29-November 3, 2023}{Ottawa, ON, Canada}
\acmBooktitle{Proceedings of the 31st ACM International Conference on Multimedia (MM '23), October 29-November 3, 2023, Ottawa, ON, Canada}

\usepackage{hyperref}
\hypersetup{
  colorlinks   = true, 
  urlcolor     = red, 
  linkcolor    = blue, 
  citecolor   = green 
}

\usepackage{amsmath}
\usepackage{booktabs}
\usepackage{caption,setspace}
\usepackage{multirow}
\usepackage{float}
\newcommand{\tabincell}[2]{\begin{tabular}{@{}#1@{}}#2\end{tabular}}  
\usepackage{balance} 

\usepackage{booktabs}
\usepackage{adjustbox}

\usepackage{bm}
\usepackage{pifont}
\usepackage{xcolor}
\usepackage{array}
\usepackage{wrapfig}

\newcommand{\cmark}{\ding{51}}%
\newcommand{\xmark}{\ding{55}}%
\newcolumntype{R}[2]{%
    >{\adjustbox{angle=#1,lap=\width-(#2)}\bgroup}%
    l%
    <{\egroup}%
}
\newcommand*\rot{\multicolumn{1}{R{35}{1em}}}
\newcolumntype{P}[1]{>{\centering\arraybackslash}p{#1}} 
\newcolumntype{M}[1]{>{\centering\arraybackslash}m{#1}} 

\usepackage[capitalize]{cleveref}
\crefname{section}{Sec.}{Secs.}
\Crefname{section}{Section}{Sections}
\Crefname{table}{Table}{Tables}
\crefname{table}{Tab.}{Tabs.}

\newcommand{\ie}{\textit{i}.\textit{e}.}
\newcommand{\eg}{\textit{e}.\textit{g}.}

\begin{document}

\title{AniPixel: Towards Animatable Pixel-Aligned Human Avatar}

\author{Jinlong Fan} 
\email{jfan0939@uni.sydney.edu.au}
\affiliation{%
  \institution{The University of Sydney}
  \city{Sydney}
  \state{NSW}
  \country{Australia}
}

\author{Jing Zhang}
\email{jing.zhang1@sydney.edu.au}
\affiliation{%
  \institution{The University of Sydney}
  \city{Sydney}
  \state{NSW}
  \country{Australia}
}

\author{Zhi Hou}
\email{zhou9878@uni.sydney.edu.au}
\affiliation{%
  \institution{The University of Sydney}
  \city{Sydney}
  \state{NSW}
  \country{Australia}
}

\author{Dacheng Tao}
\email{dacheng.tao@gmail.com}
\affiliation{%
  \institution{The University of Sydney}
  \city{Sydney}
  \state{NSW}
  \country{Australia}
}

\renewcommand{\shortauthors}{Jinlong Fan, Jing Zhang, Zhi Hou, \& Dacheng Tao}

\begin{abstract}
    Although human reconstruction typically results in human-specific avatars, recent 3D scene reconstruction techniques utilizing pixel-aligned features show promise in generalizing to new scenes. Applying these techniques to human avatar reconstruction can result in a volumetric avatar with generalizability but limited animatability due to rendering only being possible for static representations. In this paper, we propose AniPixel, a novel animatable and generalizable human avatar reconstruction method that leverages pixel-aligned features for body geometry prediction and RGB color blending. Technically, to align the canonical space with the target space and the observation space, we propose a bidirectional neural skinning field based on skeleton-driven deformation to establish the target-to-canonical and canonical-to-observation correspondences. Then, we disentangle the canonical body geometry into a normalized neutral-sized body and a subject-specific residual for better generalizability. As the geometry and appearance are closely related, we introduce pixel-aligned features to facilitate the body geometry prediction and detailed surface normals to reinforce the RGB color blending. We also devise a pose-dependent and view direction-related shading module to represent the local illumination variance. Experiments show that AniPixel renders comparable novel views while delivering better novel pose animation results than state-of-the-art methods. Code will be released at \href{https://github.com/loong8888/AniPixel}{AniPixel}.
\end{abstract}



\maketitle

\section{Introduction}
Human animation and free-view rendering have a variety of applications such as telepresence, movies, video games, and sports broadcasting~\cite{zhang2020empowering}. Conventionally, 3D human avatar reconstruction requires expensive setups such as dense multi-view camera rigs or accurate depth sensors~\cite{guoRelightablesVolumetricPerformance2019,colletHighqualityStreamableFreeviewpoint2015,douFusion4dRealtimePerformance2016}. With the recent success of neural radiance field (NeRF) representation~\cite{mildenhallNerfRepresentingScenes2021}, a line of works has tried to reconstruct volumetric avatars in radiance field~\cite{pengNeuralBodyImplicit2021,pengAnimatableNeuralRadiance2021a,wengHumannerfFreeviewpointRendering2022}. These neural avatars are human-specific and the model has to be trained from scratch for an unseen person, which is tedious and frustrating in applications. Another line of works leverages pixel-aligned features to reconstruct a generalizable avatar from sparse multi-view images~\cite{saitoPifuPixelalignedImplicit2019,saitoPifuhdMultilevelPixelaligned2020,rajPvaPixelalignedVolumetric2021,kwonNeuralHumanPerformer2021,yangS3NeuralShape2021,rajAnrArticulatedNeural2021}, which can be tested directly on unseen persons. Although pixel-aligned volumetric avatars can achieve photo-realistic novel-view synthesis for unseen persons, they are limited to $\textit{still}$ humans since the position of query points under the target pose must be the same as that of the ones used to extract the pixel-aligned features under the input pose, which prevents volumetric avatar from pose-controllable animation. 

\begin{table}
\centering
\small
\resizebox{\linewidth}{!}{%
    \begin{tabular}{cccccc|l}
    \rot{generalizable}  & \rot{animatable} & \rot{surface normals} &\rot{learnable lbs} & \rot{latent codes} & \rot{deformation} &  \\
    \hline
    \xmark & \cmark & \xmark & \xmark & per-frame & F & NeuralBody~\cite{pengNeuralBodyImplicit2021} \\
    \xmark & \cmark & \xmark & \cmark & per-frame & B & AniNeRF \cite{pengAnimatableNeuralRadiance2021a} \\
    \cmark & \cmark & \xmark & \xmark & \xmark & F+B& MPS-NeRF~\cite{gaoMPSNeRFGeneralizable3D2022} \\
    \cmark & \xmark & \xmark & \xmark & \xmark & -& KeypointNeRF~\cite{mihajlovicKeypointNeRFGeneralizingImagebased2022} \\
    \hline
    \cmark & \cmark & \cmark & \cmark &  per-idt. & F+B& \textbf{Ours} \\
\end{tabular}}
\caption{\small{\textbf{Design differences}. \textbf{F} and \textbf{B} mean the \textbf{F}orward and \textbf{B}ackward skinning, respectively. Per-frame indicates that the latent code is for each frame while per-idt means a latent code for each person. learnable lbs indicates if the blend weights are learnable.
}}
\label{tab:designdiff}
\vspace{-1cm}
\end{table}

In this work, our goal is to make the pixel-aligned human avatar animatable, whilst preserving its capability to generalize to previously unseen persons, as well as facilitating decent-quality rendering when generating novel views using sparse multi-view images. However, there is a significant challenge, i.e., how to align the dynamic points in the target space to the input space. Some recent works extend NeRF with a non-rigid deformation field to represent dynamic scenes~\cite{parkNerfiesDeformableNeural2021,pumarolaDnerfNeuralRadiance2021,tretschkNonrigidNeuralRadiance2021} but jointly learning NeRF and the deformation field is ill-posed and prone to local minima~\cite{liNeuralSceneFlow2021,pengAnimatableNeuralRadiance2021a}. For human reconstruction, skeleton motion is often taken as prior to constraining the deformation field~\cite{pengNeuralBodyImplicit2021,pengAnimatableNeuralRadiance2021a,zhaoHumanNeRFEfficientlyGenerated2022}. As these methods commonly render the body directly from the canonical space, they only focus on the deformation from observation space to canonical space. But in our setting, the pixel-aligned features are extracted in a separate input space, which makes the problem more challenging. To deal with that, we devise a bidirectional neural skinning field that enables both target-to-canonical and canonical-to-observation deformation. Meanwhile, we also leverage human priors in the learning process to optimize the deformation field~\cite{loperSMPLSkinnedMultiperson2015,lewisPoseSpaceDeformation2000a}. 

Nevertheless, the diverse body shapes and appearances among different persons pose significant challenges to appealing animatable pixel-aligned avatars. To attain better generalization capability and reconstruct more geometry details, we devise a canonical body and disentangle it into the human-shared neutral part and a subject-specific part. Humans share similar body structures, but each person also has a unique body shape, appearance, style of dress, etc. The shared part in our method is represented as a normalized neutral-sized body and the subject-specific part is described by a residual displacement field. On the other hand, for appearance reconstruction, blending weights are predicted using the fused pixel-aligned features~\cite{wangIbrnetLearningMultiview2021}. In existing methods, the blended colors are pose-independent, but in our setting, the target pose can be different from the input pose. To that end, we incorporate a shading module to predict a per-point scalar shading factor to modulate the blended colors for representing the pose-related local illumination. Combing the appearance module with the canonical geometry module, we can generate a holistic volumetric avatar.

We evaluate our Animatable Pixel-aligned human avatar dubbed AniPixel, on the Human3.6M~\cite{ionescuHuman36mLarge2013} and ZJUMoCap~\cite{pengNeuralBodyImplicit2021} datasets that provide synchronized multi-view video sequences of dynamic humans. Both for the novel view synthesis and novel pose animation, AniPixel exhibits state-of-the-art performance, and surprisingly outperforms human-specific methods on the animation task. 

In summary, our method reconstructs a volumetric human avatar that is both animatable and generalizable. Novel views of unseen persons in novel poses can be directly rendered from sparse multi-view images, which is of great practical significance in real-world applications. The contribution of this paper is three-fold.
    1) We devise a bidirectional neural skinning field and a neutralized canonical space to align the target pose with the input pose, making the pixel-aligned human avatar animatable.
    2) We represent generalizable human body geometry by disentangling it into a neutral-sized shared body and a subject-specific residual field for better generalizability.
    3) We leverage an extra shading module to modulate the RGB color to better represent the local illumination variance. Our method can render even better results in novel view and novel pose for unseen persons than human-specific methods.

\vspace{-4mm}
\section{Related Work}

\begin{figure*}[t!]
\begin{center}
\includegraphics[width=0.9\textwidth]{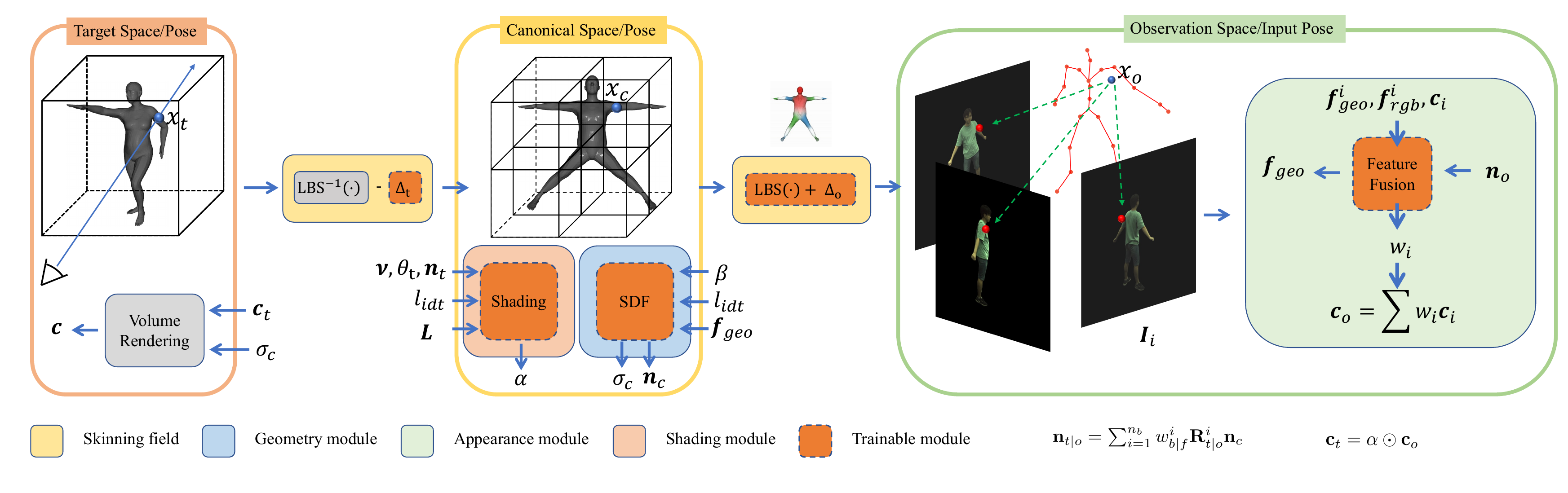}. 
\caption{\textbf{Method overview.} We propose AniPixel, an animatable and generalizable method that only takes sparse multi-view images as input. Specifically, our method includes four modules: a) \textbf{Skinning field} module outputs a bidirectional neural skinning field to align the target pose and the input pose with the canonical pose, which integrates skeleton-driven deformation with a learnable blending weight field. Given query points $\mathbf{x}_t$ in target space, they are first transformed to the canonical space as $\mathbf{x}_c$ via the backward skinning and then to the observation space as $\mathbf{x}_o$ via the forward skinning; b) \textbf{Canonical Geometry Module} represents body geometry in the canonical pose as an SDF which is shared by multiple persons. For better generalizability, we disentangle the canonical body geometry to a neutral-sized body and a pose-dependent and human-specific residual displacement field $\Delta$; c) \textbf{Appearance Module} blends RGB colors $\mathbf{c}_i$ from the input images $\mathbf{I}_i$ to colors $\mathbf{c}_o$ for points $x_o$ in the observation space; d) \textbf{Shading Module} leverages scalar shading factor $\alpha$ to modulate the output color $\mathbf{c}_o$ to represent the local illumination variance. Final RGB colors $\mathbf{c}$ according to the target points $x_t$ is accumulated through volume rendering.}
\label{fig:overview}
\end{center}
\end{figure*}

\paragraph{Neural rendering.}
Recently, various neural scene representations have been presented for novel view synthesis~\cite{lombardiNeuralVolumesLearning2019,sitzmannSceneRepresentationNetworks2019} and geometric reconstructions~\cite{parkDeepsdfLearningContinuous2019,meschederOccupancyNetworksLearning2019}. In particular, NeRF~\cite{mildenhallNerfRepresentingScenes2021} that combines MLPs with differentiable volumetric rendering achieves photo-realistic view synthesis. Standard NeRF needs per-scene optimization making it expensive for real-life applications. A bunch of following works have tried to advance it with generalizability~\cite{yuPixelnerfNeuralRadiance2021,wangIbrnetLearningMultiview2021,chenMvsnerfFastGeneralizable2021,trevithickGrfLearningGeneral2021} so that the trained model can directly synthesize novel views of novel scenes from the multi-view input images without re-training. In these works, the pixel-aligned features-based technique often plays a fundamental role. However, it can only work on static scenes, \eg, for rendering humans~\cite{saitoPifuPixelalignedImplicit2019,saitoPifuhdMultilevelPixelaligned2020,zhaoHumanNeRFEfficientlyGenerated2022,noguchiNeuralArticulatedRadiance2021}, the person is required to be still. In this paper, we devise a bidirectional neural skinning field and a neutral canonical space to make the pixel-aligned features adaptive to dynamic humans and the reconstructed volumetric avatar animatable.

\paragraph{Human animation.}
As a common approach, skeletal animation~\cite{lewisPoseSpaceDeformation2000a,kavanSkinningDualQuaternions2007} combines skeleton and per-vertex blend weight to animate a human mesh. Based on the Skinned Multi-Person Linear model (SMPL)~\cite{loperSMPLSkinnedMultiperson2015}, the human mesh can be animated with SMPL parameters fitted from images~\cite{suAnerfArticulatedNeural2021,liuNeuralActorNeural2021}. However, SMPL can only describe naked persons and can not directly render photo-realistic images. Recent works integrate SMPL with NeRF to capture human clothing and hair~\cite{wuMultiviewNeuralHuman2020,huangArchAnimatableReconstruction2020,heARCHAnimationreadyClothed2021}. To account for dynamic humans, deformation fields are devised to deform the posed body in target space to canonical space, where the density and color are predicted~\cite{pengNeuralBodyImplicit2021,pengAnimatableNeuralRadiance2021a}. To ensure stability during training, human priors are often introduced~\cite{wengHumannerfFreeviewpointRendering2022} or used to initialize the motion field~\cite{pengNeuralBodyImplicit2021,gaoMPSNeRFGeneralizable3D2022,dual-nerf}. In Dual-NeRF~\cite{dual-nerf}, the radiance field of the canonical body and the lighting conditions in the world space is represented separately using MLPs, resulting in a more accurate depiction of pose-dependent color effects. The reconstructed volumetric avatar in the canonical space can be driven by novel poses and render novel view images. However, these models are human-specific and have to be trained from scratch for each new subject. In recent research, attempts have been made to reconstruct generalizable avatars~\cite{kwonNeuralHumanPerformer2021, chengGeneralizableNeuralPerformer2022}. NHP~\cite{kwonNeuralHumanPerformer2021} employs a Temporal Transformer for aggregating skeletal features and a Multi-view Transformer for fusing time-augmented features. GNR~\cite{chengGeneralizableNeuralPerformer2022} proposes an extra occlusion-aware appearance blending module to guide the blending of appearances from source views. Similarly, our approach shares the same objective. By utilizing a multi-person shared canonical body and pixel-aligned appearance features, our method demonstrates the ability to generalize to unseen individuals. The most related work to our method is MPS-NeRF~\cite{gaoMPSNeRFGeneralizable3D2022}, which directly utilizes skeleton-driven deformation to align humans and the canonical geometry is an SMPL body. In contrast, we use learnable skinning fields to align the points and supplement a neutral-sized shared body with a residual field to depict the dynamic body. Besides, the RGB colors in MPS-NeRF are predicted using fused features, while our method outputs the colors by blending input ones, which has been proven more efficient~\cite{wangIbrnetLearningMultiview2021}. We also take into account the local illumination variance in color rendering.

\section{Method}
We propose AniPixel which can directly render realistic images of an unseen person in novel views and novel poses taking only multi-view images as input. For the input multi-view images, we assume the calibration parameters and the foreground human mask are known. We also assume the parameters of a 3D human parametric model are fitted both for the target pose and the input pose. We use SMPL as our parametric model. The 3D joints for the input pose are also regressed.

The overview of the proposed method is illustrated in Figure~\ref{fig:overview}. Following the rendering scheme of NeRF~\cite{mildenhallNerfRepresentingScenes2021,maxOpticalModelsDirect1995}, we cast rays to the target space which pass the camera center and the pixel, and then sample points along the rays. The sampled points $\mathbf{x}_t$ in target pose are first transformed to the canonical space as $\mathbf{x}_c$, and then to the observation space as $\mathbf{x}_o$ via the bidirectional skinning field which is based on skeleton-driven deformation and neural blend weight field (Section~\ref{sec:motion}). The body geometry is stored in a canonical neural field (Section~\ref{sec:canonical}) and appearance information of $\mathbf{x}_t$ is derived from input images using pixel-aligned features (Section~\ref{sec:app}). An extra shading module is leveraged to represent the local illumination variance (Section~\ref{sec:shading}). The final color $\mathbf{c}$ is accumulated through differentiable volume rendering~\cite{kajiyaRayTracingVolume1984}.

\subsection{Bidirectional Skinning Field}
\label{sec:motion}

To align the target pose with the input pose, we propose a bidirectional skinning field including a backward skinning field to transform points in the target space to the canonical space and a forward skinning field to transform the canonical points to the observation space. We use the linear blend skinning (LBS)~\cite{lewisPoseSpaceDeformation2000a} as the skinning algorithm. But the original LBS weights are only defined on SMPL mesh points. Generally, there are two ways to diffuse the weights to any 3D point. One way is to assign the weight using its nearest neighbor (NN) on the SMPL mesh or through barycentric interpolation of Top-k closest vertices~\cite{pengNeuralBodyImplicit2021,wangNeuralNovelActor2022,gaoMPSNeRFGeneralizable3D2022}. The other way is to predict the per-point weights using a neural network~\cite{pengAnimatableNeuralRadiance2021a,wengHumannerfFreeviewpointRendering2022,liuNeuralActorNeural2021}. Learnable blend weights are more flexible and can represent non-rigid deformation more precisely. However, we found that if the backward and forward blend weights are both learnable and optimized simultaneously, the training would be unstable and hard to converge. So in this work, we devise the backward weights field as a deterministic NN field and the forward one as a learnable neural field, which demonstrate adequate capacity in our experiment.

\paragraph{Backward skinning.}
Given sampling points $\mathbf{x}_t$ in the target space and the target pose parameters $\mathbf{\theta}_t$, we first calculate the bone transformations $\mathbf{B}_t = \{\mathbf{B}_t^i\}_{i=1}^{24}$ corresponding to pose $\mathbf{\theta}_t$. Each $\mathbf{B}^i$ is a $4\times4$ rotation-translation matrix. The skinning weight vector is defined as $\mathbf{w}_b \in [0, 1]^{25}$, s.t. $\sum_{i=1}^{25}{{w}_b^i} = 1$. Note that here we add an extra blend weight $w_{bg}$ for static background points~\cite{wengHumannerfFreeviewpointRendering2022}, so the skinning field can represent both the foreground and background motions. The background weights are calculated as $w_{bg} = 1 - \sum_{i=1}^{24}{w_{fg}^i}$. For each point in space, we assign the skinning weights of its NNs on the body surface, and if the closest distance is greater than a threshold we assume it is a background point and set $w_{fg}^i=0, w_{bg}=1$. For subject-specific geometry that can not be shared in the canonical body, \eg, clothes and hair, we model the residual as a displacement field and implement it as an MLP network $F_{\sigma_d}: (\phi(\mathbf{x}), l_{idt}, \mathbf{\theta}) \mapsto \Delta$, where $\sigma_d$ is the network parameters and $\Delta$ is the per-point displacement, $l_{idt}$ is the per-identity latent code and $\phi(\cdot)$ is the position encoding function~\cite{tancikFourierFeaturesLet2020}. Combining the skinning field with the displacement field, we can transform points $\mathbf{x}_t$ as:
\begin{equation}
\begin{split}
    \mathbf{x}_c &= LBS(\mathbf{x}_t)^{-1} - \mathbf{\Delta}_t \\
                 &= (\sum_{i=1}^{n_b}{w_b^i(\mathbf{x}_t, \mathbf{S}_t) \cdot \mathbf{B}_t^i(\mathbf{\theta}_t, \mathbf{\theta}_c))^{-1} \cdot \mathbf{x}_t} - \mathbf{\Delta}_t, 
\end{split}
\end{equation}
where $\mathbf{S}_t$ is the target SMPL parameters, $n_b = 25$, $\mathbf{\theta}_c$ denotes the canonical pose parameters, and $\Delta_t$ is the displacement regarding to target pose $\mathbf{\theta}_t$.

\paragraph{Implicit forward skinning.}
The pixel-aligned features are sampled by projecting 3D points to the 2D feature maps, so the results are sensitive to the position of 3D points. In order to align the canonical points with the corresponding ones in the observation space more precisely, we use a neural blend weight field in forward skinning. As this weight field is multi-person shared, we also condition it on $l_{idt}$. We implement it as a separate MLP network $F_{\sigma_w}:(\phi(x_c), l_{idt}) \mapsto \mathbf{w}_{f}$. However, the weight field ($\mathbf{w}_{f}$) is under-constrained, and jointly learning with the neural field is prone to local minima~\cite{liNeuralSceneFlow2021,pengAnimatableNeuralRadiance2021a}. To deal with that, we initialize the blend weight field with the pre-defined values $\mathbf{w}_{init}$ in the SMPL model. The forward skinning weights are generated via the softmax activation function:
\begin{equation}
\resizebox{0.9\hsize}{!}{$\mathbf{w}_f(\mathbf{x}_c) = softmax(F_{\sigma_w}(\phi(\mathbf{x}_c), l_{idt}) + log(\mathbf{w}_{init}(\mathbf{x}_c))).$}
\end{equation}

When transforming points from the canonical space into the observation space, we need to add back the human-specific displacement:
\begin{equation}
\begin{split}
    \mathbf{x}_o &= LBS(\mathbf{x}_c) + \mathbf{\Delta}_o \\
                 &=\sum_{i=1}^{n_b}{w_f^i(\mathbf{x}_c) \cdot \mathbf{B}_o^i(\mathbf{\theta}_c, \mathbf{\theta}_o) \cdot \mathbf{x}_c} + \mathbf{\Delta}_o,
\end{split}
\end{equation}
where $\Delta_o$ is the displacement regarding to input pose $\mathbf{\theta}_o$.

\begin{figure*}[t!]
\begin{center}
\includegraphics[width=.9\linewidth]{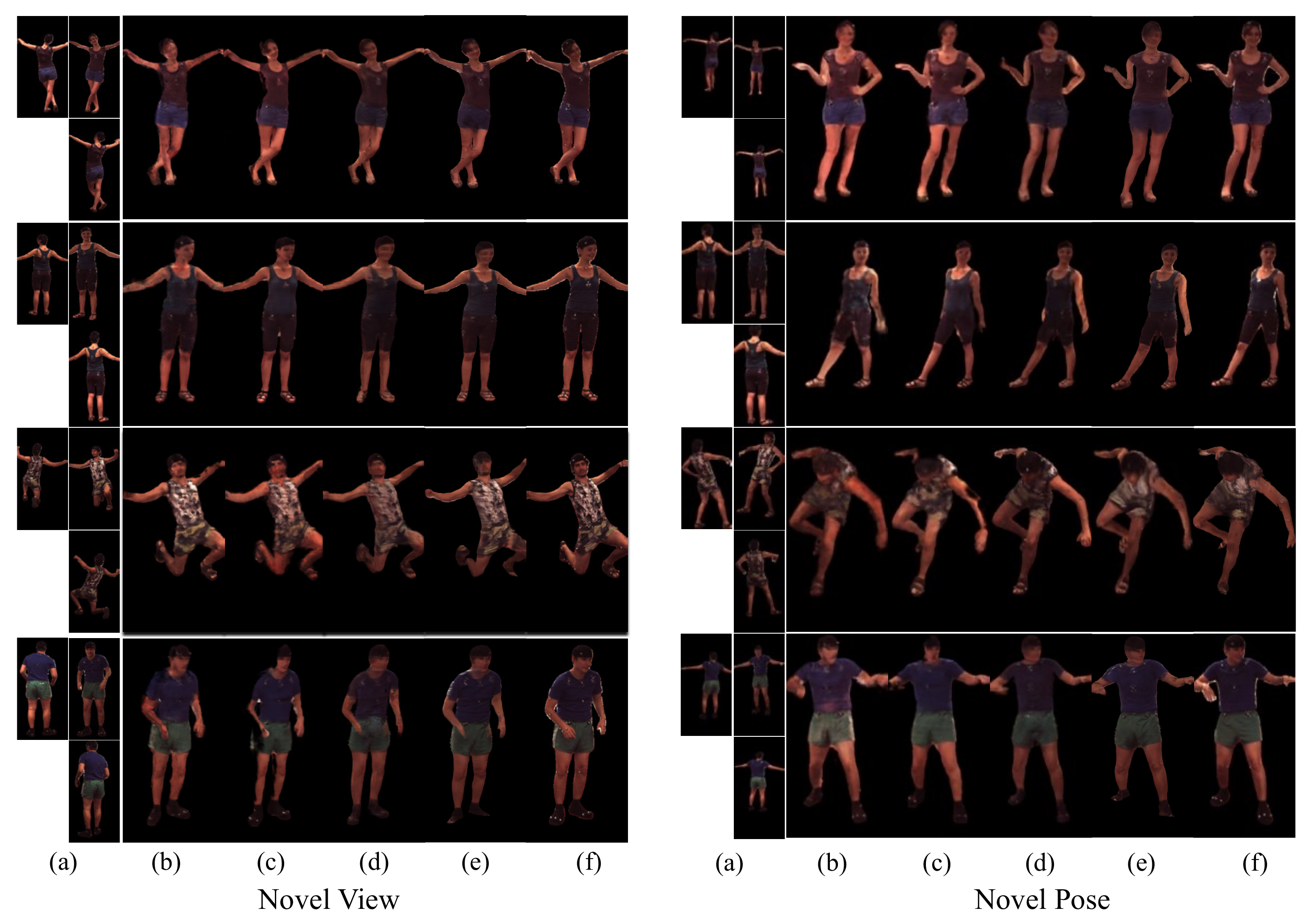}
\caption{Novel view and novel pose rendering results on the Human3.6M dataset. For each part, there are (a) the input three views and results of (b) AniNeRF~\protect\cite{pengAnimatableNeuralRadiance2021a}, (c) NeuralBody~\protect\cite{pengNeuralBodyImplicit2021}, (d) MPS-NeRF~\protect\cite{gaoMPSNeRFGeneralizable3D2022}, (e) our method, and (f) the ground truth. NeuralBody and AniNeRF are human-specific methods and do not need multiple input views in testing. The results in the above two settings are rendered given the target camera parameters and novel pose parameters, respectively.}
\label{fig:h36m}
\end{center}
\end{figure*}

\subsection{Canonical Geometry Module}
\label{sec:canonical}

For a specific person, we can disentangle its body geometry into a multi-person shared part and a residual part. The latter is modeled as the displacement field $\Delta$ and the former is stored in a canonical neural field. When shared by multiple persons, the canonical field should have the ability to distinguish the geometry difference between each subject, \eg, bone length. Utilizing the shape parameters $\mathbf{\beta}$ from the SMPL model, we can represent the body shape approximately, \eg, fat or slim. More than that, we condition the neural field on per-identity latent code $l_{idt}$ for the other variations, \eg, the shape of shoes. To make learning easier, we also resize the body of each person to align with a pre-defined neutral-sized canonical body via the minimum 3D bounding box of the SMPL mesh. The pixel-aligned features from the observation space can provide significant clues for geometry prediction, so we also condition the neural field on the fused pixel-aligned features. The canonical neural field is defined as a signed distance field (SDF)~\cite{yarivVolumeRenderingNeural2021} and HashGrid~\cite{mullerInstantNeuralGraphics2022} is taken as the position encoding method. An MLP is used to predict the SDF: $F_{\sigma_s}: (\phi(\mathbf{x}_c), \mathbf{\beta}, l_{idt}, \mathbf{f}_{geo}) \mapsto s$, where $\mathbf{f}_{geo}$ is the geometry features. Given a point $\mathbf{x}_c$ in the canonical space, its SDF is:
\begin{equation}
    s = F_{\sigma_s}(\phi(\mathbf{x}_c), \mathbf{\beta}, l_{idt}, \mathbf{f}_{geo}),
\end{equation}
where $\phi(\cdot)$ is the HashGrid position encoding function.

Given the SDF, surface normals $\mathbf{n}_c$ can be calculated as $\mathbf{n}_c = \nabla F_{\sigma_s} / ||\nabla F_{\sigma_s}||_2$, where gradient $\nabla F$ can be obtained by network backpropagation. And the surface normal vectors are transformed to the target and observation space by using the rotational part of the backward skinning and forward skinning respectively: $\mathbf{n}_{t|o} = \sum_{i=1}^{n_b}w_{b|f}^i\mathbf{R}_{t|o}^i\mathbf{n}_c$. Following the SDF-based volume rendering formulation~\cite{yarivVolumeRenderingNeural2021}, we convert SDF values into density values $\sigma$ using the scaled CDF of the Laplace distribution:
\begin{equation}
    \sigma = \frac{1}{b} (\frac{1}{2} + \frac{1}{2} sign(s) (exp(-\frac{|s|}{b}) - 1),
\end{equation}
where $b$ is a learnable parameter.

\begin{table*}[t!]
	\centering
        \small
	\vspace{-7pt}
	\setlength{\tabcolsep}{1.2 mm}{
				\begin{tabular}{c|cccccccc|cccccccc}
					\toprule
					\multicolumn{1}{c|}{\multirow{3}{*} {Subject}} & \multicolumn{8}{c|}{Novel View Synthesis} & \multicolumn{8}{c}{Novel Pose Synthesis} \\
					& \multicolumn{4}{c}{PSNR} & \multicolumn{4}{c|}{SSIM} & \multicolumn{4}{c}{PSNR} & \multicolumn{4}{c}{SSIM} \\
					\cline{2-17}
					& ~~~NB~~~ & \!\!AniNeRF\!\!  & ~MPS~ & ~Ours~ & ~~~NB~~~ & \!\!AniNeRF\!\!  & ~MPS~ & ~Ours~ & ~~~NB~~~ & \!\!AniNeRF\!\!  & ~MPS~ & ~Ours~ & ~~~NB~~~ & \!\!AniNeRF\!\!  & ~MPS~ &~Ours~ \\
					\hline
					S1    & 22.87  & 22.05 & \textbf{25.40} & \underline{25.15} & 0.897 & 0.888 & \textbf{0.926} & \underline{0.912} & \underline{22.11} & 21.37 & 21.87 & \textbf{24.60} & 0.879 &0.868	& \underline{0.880} & \textbf{0.896} \\
					S5    & \underline{24.60}  & 23.27 & 24.30 & \textbf{24.63}  & \textbf{0.917} & 0.892 & \underline{0.908}& 0.894 & \underline{23.51} & 22.29 & 21.49& \textbf{24.19} & \textbf{0.897} & 0.875 & 0.871& \underline{0.883} \\
					S6    & 22.82 & 21.13 & \underline{23.94}& \textbf{24.49} & 0.888 & 0.854 & \underline{0.893} & \textbf{0.897} & 23.52 & 22.59 & \underline{23.63}& \textbf{23.85} & 0.889 & 0.884 & \underline{0.891}& \textbf{0.894} \\
					S7    & 23.17 & 22.50 & \underline{24.27}& \textbf{24.60} & \textbf{0.914} & 0.890 & \underline{0.911}& 0.901 & \underline{22.33} & 22.22 & 21.88& \textbf{23.87} & \textbf{0.889} & 0.878 & 0.868& \underline{0.884} \\
					S8    & 21.72 & 22.75 & \underline{23.66}& \textbf{24.41} & 0.894 & \underline{0.898} & \textbf{0.920}& 0.895 & 20.94 & \underline{21.78} & 21.15 & \textbf{24.03} & 0.876 & 0.882 & \underline{0.888}& \textbf{0.892} \\
					S9    & 24.28 & \underline{24.72} & 24.55 & \textbf{25.87} & \underline{0.910} & 0.908 & 0.899& \textbf{0.918} & 23.04 & \underline{23.72} & 23.33 & \textbf{24.83} & 0.884 & \underline{0.886} & 0.875 & \textbf{0.904} \\
					S11   & 23.70 & 24.55 & \textbf{25.12}& \underline{24.95} & 0.896 & 0.902 & \textbf{0.913}& \underline{0.905} & 23.72 & \underline{23.91} & 23.53& \textbf{24.06} & 0.884 & 0.889 & \underline{0.891}& \textbf{0.895} \\
					Average & 23.31 & 23.00    & \underline{24.06}& \textbf{24.87} & \underline{0.903} & 0.890  & \textbf{0.910}& \underline{0.903} & \underline{22.74} & 22.55 & 22.41& \textbf{24.20} & \underline{0.885} & 0.880 & 0.881 & \textbf{0.893} \\
					
					\bottomrule
				\end{tabular}%
		}
        \caption{Comparison of our method with \textbf{N}eural\textbf{B}ody~\protect\cite{pengNeuralBodyImplicit2021}, AniNeRF~\protect\cite{pengAnimatableNeuralRadiance2021a}, \textbf{MPS}-NeRF~\protect\cite{gaoMPSNeRFGeneralizable3D2022} on the Human3.6M dataset. The best and second-best results are marked in bold and underlined, respectively.}
        \label{tab:h36m}
        \vspace{-0.25cm}
\end{table*}

\subsection{Appearance Module}
\label{sec:app}

Following~\cite{wangIbrnetLearningMultiview2021}, we utilize pixel-aligned features from the sparse multi-view input images in the observation space to blend RGB colors, making the appearance module generalizable and requiring only sparse input views.

Given the transformed 3D query points $\mathbf{x}_o$, we project them onto the input images ${\mathbf{I}_i}$ and the extracted feature maps ${\mathbf{F}_i}$ by perspective projection $\Pi(\mathbf{x}_o|\mathbf{P}_i)$. The pixel-aligned colors $\mathbf{c}_i$ and features $\mathbf{f}_i$ are sampled through bilinear interpolation. Considering the camera rays in the target space are bent after being transformed to the observation space, unlike~\cite{mihajlovicKeypointNeRFGeneralizingImagebased2022}, we do not directly take the target view directions as input to represent the view-dependent effects. Instead, we introduce the detailed surface normals to better indicate the per-point directions and describe the view-related effects as a shading factor (Sec. \ref{sec:shading}). Similar to~\cite{wangIbrnetLearningMultiview2021}, we output the RGB color $\mathbf{c}_o$ as a weighted sum of all the input colors $\mathbf{c}_i$, and the blending weights for each input view are predicted using a feature fusion function $F_{\sigma_c}: (\phi^{\prime}(\mathbf{x}_o, \mathbf{J}_o), \mathbf{f}_{geo}^i, \mathbf{f}_{rgb}^i, \mathbf{c}_i, \mathbf{n}_o) \mapsto (w_i, \mathbf{f}_{geo})$, where $\mathbf{f}_{geo}^i$ and $\mathbf{f}_{rgb}^i$ are the input geometry and appearance features for $i$th view, $\phi^{\prime}(\cdot)$ is relative spatial encoding function~\cite{mihajlovicKeypointNeRFGeneralizingImagebased2022}, $w_i$ is the output blending weights for $i$th view, s.t. $w_i \in [0,1]$ and $\sum{w_i}=1$. We also output the fused geometry features $\mathbf{f}_{geo}$ to facilitate the canonical geometry reconstruction. Please see the supplemental material for further architectural details. The blended color $\mathbf{c}_o$ can be written as:
\begin{equation}
    \mathbf{c}_o = \sum{w_i\mathbf{c}_i} = \sum{F_{\sigma_c}(\mathbf{f}_{geo}^i, \mathbf{f}_{rgb}^i, \mathbf{c}_i, \mathbf{n}_o)\mathbf{c}_i}.
\end{equation}

\subsection{Shading Module}
\label{sec:shading}
As the appearance information derived from the observation space is in the input pose, which is different from the target pose. When the human pose changes, local illumination on the body surface may also change, \eg, if lifting arms to the head, there might be shading on the face. Besides, the colors in the observation space are not conditioned on the target view directions $\mathbf{v}$, so the output colors $\mathbf{c}_o$ are not view-dependent, which is undesirable for realistic rendering. To tackle this problem, we devise a shading module predicting a pose-dependent and view direction-related per-point scalar shading factor to modulate the output colors $\mathbf{c}_o$. To capture subject-specific effects, \eg, different reflectance caused by variant clothes materials, we also take the identity latent code $l_{idt}$ as conditional input. Additionally, during the data collection process, the captured images are obtained over a period of time, during which environmental illumination and camera settings may change. Inspired by \cite{alldieckPhotorealisticMonocular3D2022}, we extract a global feature $\mathbf{L}$ from the input images to account for the overall per-image illumination variance. In a nuts shell, the shading module aims to accurately represent the per-image, subject-specific, pose-dependent, and view-dependent color factors. We use a shallow MLP to predict the shading factor $F_{\sigma_l}: (\mathbf{v}, \mathbf{\theta}_t, \mathbf{n}_t, l_{idt}, \mathbf{L}) \mapsto \alpha$. The modulated color is:

\begin{equation}
    \mathbf{c}_t = \alpha \odot \mathbf{c}_o = F_{\sigma_l}(\mathbf{v}, \mathbf{\theta}_t, \mathbf{n}_t, l_{idt}, \mathbf{L}) \odot \mathbf{c}_o,
\end{equation}
where $\odot$ means per-point multiplication. Through this shading factor modulation, the appearance information in observation space is adapted to the target space. Given the density and RGB color for each query point on a ray $\mathbf{r} = (\mathbf{c}_t, \sigma)$, we render the pixel color using standard volume rendering~\cite{mildenhallNerfRepresentingScenes2021}.

\begin{table}[t!]
	\centering
	\small
	\setlength{\tabcolsep}{2.5mm}{
				\begin{tabular}{rcccc}
					\toprule
					\multicolumn{1}{c}{\multirow{2}{*} {Method}} & \multicolumn{2}{c}{Novel View} & \multicolumn{2}{c}{Novel Pose} \\
					\cmidrule{2-5}    & PSNR & SSIM  & PSNR & SSIM  \\
					\midrule
                        Human-specific:&&&& \\
					NeuralBody    & \textbf{28.90} & \textbf{0.967} & 23.06 & 0.879 \\
					AniNeRF & 27.10  & 0.949 & 23.16 & 0.893  \\
					\midrule
                        Generalizable: &&&& \\
                        KeypointNeRF & 25.03  & 0.896 & N/A & N/A  \\
					Ours  & 26.39 & 0.911 & \textbf{24.87} & \textbf{0.895} \\
					\bottomrule
				\end{tabular}%
 }
        \caption{Comparison of NeuralBody~\protect\cite{pengNeuralBodyImplicit2021}, AniNeRF~\protect\cite{pengAnimatableNeuralRadiance2021a}, KeypointNeRF~\protect\cite{mihajlovicKeypointNeRFGeneralizingImagebased2022}, and our method on ZJUMoCAP. NeuralBody and AniNeRF are human-specific methods while KepointNeRF and our method are generalizable. KeypointNeRF requires the target pose and input pose to be the same and is not applicable to the novel pose setting.}
        \label{tab:zju}%
        \vspace{-1.0cm}
\end{table}

\begin{figure*}[t!]
\begin{center}
\includegraphics[width=0.9\linewidth]{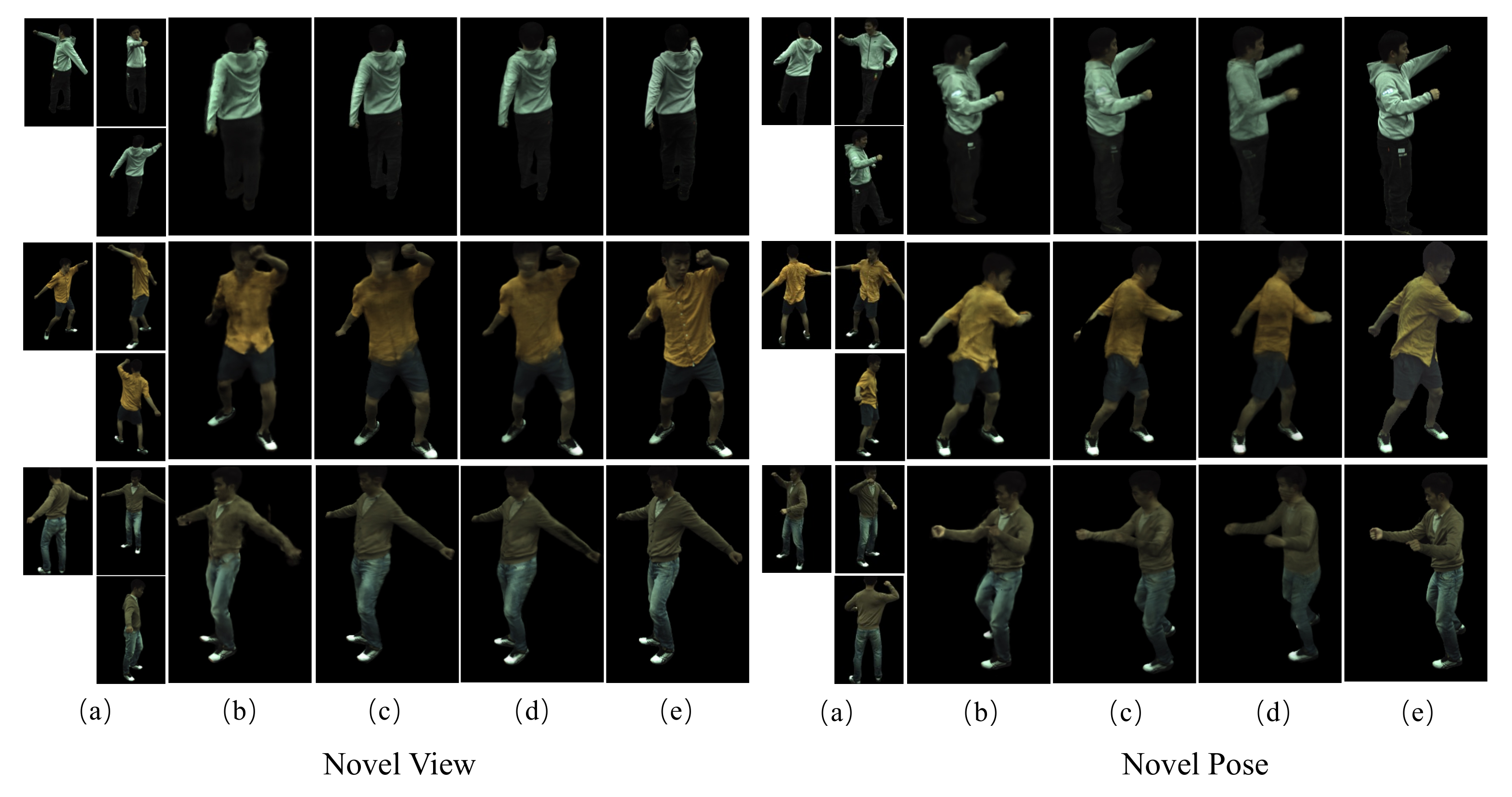}
\caption{Novel view and novel pose synthesis results on the ZJUMoCAP dataset. For each part, there are (a) the input three views and results of (b) NeuralBody~\protect\cite{pengNeuralBodyImplicit2021}, (c) KeypointNeRF~\protect\cite{mihajlovicKeypointNeRFGeneralizingImagebased2022}, (d) our methods, and (e) the ground truth. NeuralBody and AniNeRF are human-specific methods. The results in the above two settings are rendered given the target camera parameters and novel pose parameters, respectively. Since KepointNeRF is not animatable, we set the input pose to be the same as the target one during testing.}
\label{fig:zju}
\end{center}
\vspace{-0.25cm}
\end{figure*}

\subsection{Training Loss}
Our method is end-to-end trainable and all the modules are optimized jointly. The training loss is defined as:
\begin{equation}
    \mathcal{L} = \lambda_C \mathcal{L}_{C} + \lambda_M \mathcal{L}_{M} +  \lambda_N \mathcal{L}_{N} +  \lambda_S \mathcal{L}_{S} + \lambda_D  \mathcal{L}_{D}.
\end{equation}
where $\mathcal{L}_{C}$ is the color loss. Specifically, given the ground truth target image and predicted one, we apply $l_1$-distance loss and VGG perception loss~\cite{simonyanVeryDeepConvolutional2015} to supervise the training. $\mathcal{L}_{M}$ is the mask loss. We predict two versions of masks, one is rendered by volume density accumulation, and the other is generated by minimum SDF rendering as in~\cite{pengAnimatableNeuralRadiance2021a}. $\mathcal{L}_{N}$ is surface normal regularization, including smoothness loss and shape loss defined the same as in~\cite{gaoMPSNeRFGeneralizable3D2022}. $\mathcal{L}_S$ is the Eikonal loss to make sure the neural field is prone to be an SDF. The gradients of the neural fields are calculated conveniently using the automatic differentiation tool in PyTorch~\cite{paszke2017automatic}. $\mathcal{L}_{D}$ is $l_2$-norm regularization of the displacement fields, including both the forward and backward skinning fields, encouraging the residual displacement to be as small as possible. $\lambda_C$, $\lambda_M$, $\lambda_N$, $\lambda_S$ and $\lambda_D$ are loss weights to balance these loss terms. Please refer to the supplement for their settings.

\section{Experimental Results}

\paragraph{Implemental details.}
We use the Adam optimizer~\cite{kingmaAdamMethodStochastic2015} with a learning rate of $5e^{-4}$ and a batch size of 1 to train the network. To initialize the SDF, we first train 30K iterations with only the $\mathcal{L}_M$ loss and then train another 120K iterations with all the losses. In order to use VGG loss to capture high-frequency details, we render patches instead of random rays~\cite{mihajlovicKeypointNeRFGeneralizingImagebased2022}. The center of the patch is randomly sampled in the minimum bounding rectangle area of the foreground mask and the query points are sampled from the 3D bounding box derived from the SMPL model. Along each ray, 64 points are sampled for coarse rendering and 16 points for fine rendering. $l_{idt}$ is selected as the nearest person for testing. We use four Nvidia Tesla V100 GPUs for training and it takes about one day to converge. More details are provided in the supplementary material.

\paragraph{Datasets.}
We mainly evaluate our method and compare it with other methods on two public datasets. The first one is the Human3.6M dataset~\cite{ionescuHuman36mLarge2013}, which contains 4-view sequences of different actors. Following~\cite{pengAnimatableNeuralRadiance2021a,gaoMPSNeRFGeneralizable3D2022}, we conduct experiments on 7 subjects: S1, S5, S6, S7, S8, S9 and S11. We test our method using the same setting as~\cite{gaoMPSNeRFGeneralizable3D2022} for a fair comparison. The second dataset is the ZJUMocap dataset~\cite{pengNeuralBodyImplicit2021}, which provides video sequences of 10 subjects captured from 23 synchronized cameras. The splitting of training and test set is the same as~\cite{mihajlovicKeypointNeRFGeneralizingImagebased2022}, 

\begin{figure*}[t!]
\begin{center}
\includegraphics[width=.85\textwidth]{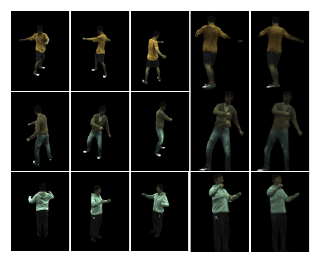}
\caption{Comparison with KeypointNeRF~\cite{mihajlovicKeypointNeRFGeneralizingImagebased2022} on novel view synthesis. First three columns are input views. The fourth column is the result of KeypointNeRF and the fifth column is the result of our AniPixel. Input poses are the same as target ones following KeypointNeRF's setting.}
\label{fig:add results zju}
\end{center}
\end{figure*}
 
\paragraph{Evaluation metrics.}
We use PSNR and SSIM metrics for quantitative evaluation. Instead of directly calculating PSNR and SSIM for the whole image, we follow previous methods~\cite{pengAnimatableNeuralRadiance2021a,pengNeuralBodyImplicit2021,gaoMPSNeRFGeneralizable3D2022} to project the 3D bounding box of the fitted SMPL mesh onto the image plane to obtain a 2D mask and only calculate PSNR and SSIM in the masked region.

\subsection{Comparison with previous methods}
\paragraph{Baselines.} 
We compare our method with recent two animatable methods, NeuralBody~\cite{pengNeuralBodyImplicit2021} and AniNeRF~\cite{pengAnimatableNeuralRadiance2021a}, and two generalizable methods, KeypointNeRF~\cite{mihajlovicKeypointNeRFGeneralizingImagebased2022} and MPS-NeRF~\cite{gaoMPSNeRFGeneralizable3D2022}. NeuralBody and AniNeRF are human-specific models that require training a single model for each subject. In evaluation, camera parameters and pose parameters are used to animate the learned neural field. KeypointNeRF and MPS-NeRF can generalize to unseen persons taking multi-view images as input. But KepointNeRF only works in static scenes and is not applicable to animation tasks.

 \begin{figure}[t!]
\begin{center}
\includegraphics[width=0.9\linewidth]{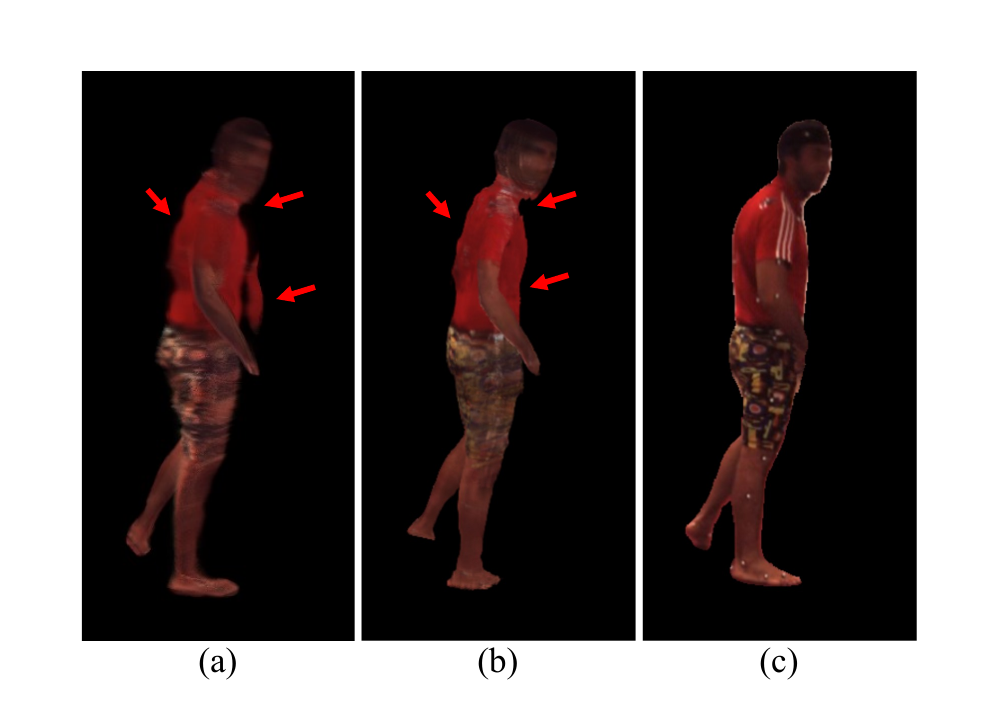}
\caption{Visual results of (a) MPS-NeRF and (b) our AniPixel as well as (c) the ground truth.}
\label{fig:diff}
\end{center}
\end{figure}

\paragraph{Results on Human3.6M}
Table~\ref{tab:h36m} presents the quantitative comparison of our AniPixel with the other three methods. For novel view synthesis, our method has marginally higher PSNR on average and comparable SSIM with other methods, which confirms the valid reconstruction of the canonical body geometry in our method. In the animation task, both our AniPixel and MPS-NeRF are tested on unseen persons. NeuralBody and AniNeRF are trained and tested on the same person. Our method outperforms all three methods both in PSNR and SSIM. Note that since AniPixel is both animatable and generalizable, it obtains about 2dB higher PSNR on average than MPS-NeRF, which we attribute to the effectiveness of the proposed neural skinning field and residual displacement field. 

Visual results are shown in Figure~\ref{fig:h36m}. It can be observed that our method can render competitive results both on novel views and unseen persons in novel poses. Our results of S6 in novel view synthesis (\ie, third row in left part) and S1 in novel pose synthesis (\ie, first row in right part) show better pixel lightness than MPS-NeRF, owing to the proposed shading module for modeling local illumination variance. Based on the bidirectional skinning field, our method can also align the poses more precisely as demonstrated in Figure~\ref{fig:diff}. Video demo and more visual results are included in the supplementary material.

\paragraph{Results on ZJUMoCap.} 
The quantitative results and rendering results compared with NeuralBody and KeypointNeRF are listed in Table~\ref{tab:zju} and shown in Figure~\ref{fig:zju} respectively. Not that NeuralBody is a human-specific method and takes target camera parameters to render novel view images and pose parameters to synthesize novel pose images. KeypointNeRF is only applicable to static humans and the input pose should be selected to be the same as the target one, while our method takes a different pose from the target pose as input. Even under a more challenging setting, our AniPixel can still output comparable visual results on novel view synthesis and novel pose synthesis with other methods and even obtains higher objective results (\eg, about 1.8dB higher PSNR) on novel pose synthesis than the human-specific methods.

\paragraph{Self-occlusion artifacts.}
In KeypointNeRF~\cite{mihajlovicKeypointNeRFGeneralizingImagebased2022}, for each pixel in the target image, the pixel color is calculated as the blending of input ones, and the blending weights are predicted based on pixel-aligned features. However, all sampled query points on one target ray share the same pixel-aligned features. The relative spatial encoding of each point may result in varying weights between them. But the slightly different weights in empty space still have a chance to output floating artifacts in self-occlusion areas. Some examples are shown in Figure~\ref{fig:add_results_zju}. 
In contrast, thanks to the explicit geometry reconstruction in canonical space, our model has the ability to distinguish empty space from the human body, which can remove floating artifacts effectively, as shown in the last column in Figure~\ref{fig:add_results_zju}.

\subsection{Ablation studies}
We conduct ablation studies on the S9 subject from the Human3.6M dataset. Results are listed in Table~\ref{tab:ablation}. When conducting ablation studies on the shading module and residual displacement fields, we simply remove them from the model. For geometry features and human identity latent code, we replace them with constant values. To verify the effectiveness of utilizing surface normals to reinforce the RGB color blending, we replace the normals with target view directions. For the learnable skinning field, it is replaced with the standard skeleton motion.

For novel view synthesis, the shading module plays an important role and the test metrics drop a lot without it. The detailed surface normals transformed from canonical space indeed benefit the color blending in observation space and including it in the model delivers higher metrics. For novel pose synthesis, geometry features show the most important impact which indicates that appearance information could be a valuable clue for geometry reconstruction. The identity latent code and displacement field benefit novel pose synthesis more than the novel view synthesis. And the identity latent code can further promote the results for both tasks.

\begin{table}[t!]
	\centering
	\small
	\setlength{\tabcolsep}{1.5mm}{
				\begin{tabular}{rcccc}
					\toprule	\multicolumn{1}{c}{\multirow{2}{*} {}} & \multicolumn{2}{c}{Novel View} & \multicolumn{2}{c}{\tabincell{c}{Novel Pose}} \\
					\cmidrule{2-5}    & PSNR & SSIM  & PSNR & SSIM  \\
					\midrule
                        w/o geo. feats &24.07&0.870&22.33&0.832 \\
                        w/o normals &23.98&0.881&22.72&0.845 \\
                        { w/o shading} & { 23.88} & { 0.889}  & { 22.82} & { 0.884} \\
                        w/o displacement & 24.56 & 0.879 & 23.21 & 0.848 \\
                        w/o identity & 24.54 & 0.875 & 23.34 & 0.863 \\
                        { w/o  learnable skinning} & { 24.29} & { 0.886}  & { 24.01} & { 0.847} \\
					\midrule
					Our AniPixel & \textbf{25.81} & \textbf{0.902} & \textbf{24.83} & \textbf{0.895} \\
					\bottomrule
				\end{tabular}%
		}
	
        \caption{Ablation study of the design choices in our model.}
        \label{tab:ablation}
        \vspace{-0.1cm}
	\end{table}

\subsection{Canonical geometry}
\begin{figure}[t!]
\begin{center}
\includegraphics[width=0.9\linewidth]{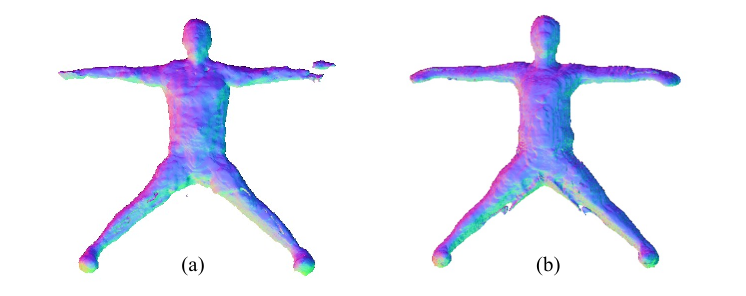}
\caption{Geometry of (a) MPS-NeRF and (b) our AniPixel in the canonical space.}
\label{fig:geo}
\end{center}
\vspace{-0.1cm}
\end{figure}

In order to validate the effectiveness of our canonical SDF, we visualize the learned body geometry of both our method and MPS-NeRF in the canonical space, as shown in Figure~\ref{fig:geo}. Compared with MPS-NeRF, the 3D canonical geometry reconstructed by our method is more complete on arms. It is because MPS-NeRF only depends on the SMPL model to constrain the shape reconstruction, which usually can not well fit the hands. In contrast, our AniPixel can reconstruct the holistic body geometry well based on the residual displacement field and the learnable skinning field.

\section{Conclusion}
We proposed an animatable and generalizable volumetric human avatar reconstruction method that could render novel views and novel poses for unseen persons from sparse multi-view images. Specifically, we devise a bidirectional neural skinning field and a neutralized canonical space to bridge the target pose and the input pose. Meanwhile, a shading module is introduced to improve the local illumination variance representation. Experiments on the Human3.6M and ZJUMoCap datasets demonstrate that the proposed approach achieves state-of-the-art performance on novel view synthesis and novel pose synthesis, and even outperforms human-specific methods on animation tasks. 

\noindent\textbf{Acknowledgement}: This work was supported by ARC FL-170100117 and IH-180100002.

\bibliographystyle{ACM-Reference-Format}  
\balance
\bibliography{mm23}


\begin{thebibliography}{51}


\ifx \showCODEN    \undefined \def \showCODEN     #1{\unskip}     \fi
\ifx \showDOI      \undefined \def \showDOI       #1{#1}\fi
\ifx \showISBNx    \undefined \def \showISBNx     #1{\unskip}     \fi
\ifx \showISBNxiii \undefined \def \showISBNxiii  #1{\unskip}     \fi
\ifx \showISSN     \undefined \def \showISSN      #1{\unskip}     \fi
\ifx \showLCCN     \undefined \def \showLCCN      #1{\unskip}     \fi
\ifx \shownote     \undefined \def \shownote      #1{#1}          \fi
\ifx \showarticletitle \undefined \def \showarticletitle #1{#1}   \fi
\ifx \showURL      \undefined \def \showURL       {\relax}        \fi
\providecommand\bibfield[2]{#2}
\providecommand\bibinfo[2]{#2}
\providecommand\natexlab[1]{#1}
\providecommand\showeprint[2][]{arXiv:#2}

\bibitem[Alldieck et~al\mbox{.}(2022)]%
        {alldieckPhotorealisticMonocular3D2022}
\bibfield{author}{\bibinfo{person}{Thiemo Alldieck}, \bibinfo{person}{Mihai
  Zanfir}, {and} \bibinfo{person}{Cristian Sminchisescu}.}
  \bibinfo{year}{2022}\natexlab{}.
\newblock \showarticletitle{Photorealistic {{Monocular 3D Reconstruction}} of
  {{Humans Wearing Clothing}}}. In \bibinfo{booktitle}{\emph{CVPR}}.
  \bibinfo{pages}{1506--1515}.
\newblock


\bibitem[Chen et~al\mbox{.}(2021)]%
        {chenMvsnerfFastGeneralizable2021}
\bibfield{author}{\bibinfo{person}{Anpei Chen}, \bibinfo{person}{Zexiang Xu},
  \bibinfo{person}{Fuqiang Zhao}, \bibinfo{person}{Xiaoshuai Zhang},
  \bibinfo{person}{Fanbo Xiang}, \bibinfo{person}{Jingyi Yu}, {and}
  \bibinfo{person}{Hao Su}.} \bibinfo{year}{2021}\natexlab{}.
\newblock \showarticletitle{Mvsnerf: {{Fast}} Generalizable Radiance Field
  Reconstruction from Multi-View Stereo}. In \bibinfo{booktitle}{\emph{ICCV}}.
  \bibinfo{pages}{14124--14133}.
\newblock


\bibitem[Cheng et~al\mbox{.}(2022)]%
        {chengGeneralizableNeuralPerformer2022}
\bibfield{author}{\bibinfo{person}{Wei Cheng}, \bibinfo{person}{Su Xu},
  \bibinfo{person}{Jingtan Piao}, \bibinfo{person}{Chen Qian},
  \bibinfo{person}{Wayne Wu}, \bibinfo{person}{Kwan-Yee Lin}, {and}
  \bibinfo{person}{Hongsheng Li}.} \bibinfo{year}{2022}\natexlab{}.
\newblock \showarticletitle{Generalizable {{Neural Performer}}: {{Learning
  Robust Radiance Fields}} for {{Human Novel View Synthesis}}}.
\newblock \bibinfo{journal}{\emph{arXiv preprint arXiv:2204.11798}}
  (\bibinfo{year}{2022}).
\newblock
\showeprint[arxiv]{2204.11798}


\bibitem[Collet et~al\mbox{.}(2015)]%
        {colletHighqualityStreamableFreeviewpoint2015}
\bibfield{author}{\bibinfo{person}{Alvaro Collet}, \bibinfo{person}{Ming
  Chuang}, \bibinfo{person}{Pat Sweeney}, \bibinfo{person}{Don Gillett},
  \bibinfo{person}{Dennis Evseev}, \bibinfo{person}{David Calabrese},
  \bibinfo{person}{Hugues Hoppe}, \bibinfo{person}{Adam Kirk}, {and}
  \bibinfo{person}{Steve Sullivan}.} \bibinfo{year}{2015}\natexlab{}.
\newblock \showarticletitle{High-Quality Streamable Free-Viewpoint Video}.
\newblock \bibinfo{journal}{\emph{ACM Transactions on Graphics (ToG)}}
  \bibinfo{volume}{34}, \bibinfo{number}{4} (\bibinfo{year}{2015}),
  \bibinfo{pages}{1--13}.
\newblock


\bibitem[Dou et~al\mbox{.}(2016)]%
        {douFusion4dRealtimePerformance2016}
\bibfield{author}{\bibinfo{person}{Mingsong Dou}, \bibinfo{person}{Sameh
  Khamis}, \bibinfo{person}{Yury Degtyarev}, \bibinfo{person}{Philip Davidson},
  \bibinfo{person}{Sean~Ryan Fanello}, \bibinfo{person}{Adarsh Kowdle},
  \bibinfo{person}{Sergio~Orts Escolano}, \bibinfo{person}{Christoph Rhemann},
  \bibinfo{person}{David Kim}, {and} \bibinfo{person}{Jonathan Taylor}.}
  \bibinfo{year}{2016}\natexlab{}.
\newblock \showarticletitle{Fusion4d: {{Real-time}} Performance Capture of
  Challenging Scenes}.
\newblock \bibinfo{journal}{\emph{ACM Transactions on Graphics (ToG)}}
  \bibinfo{volume}{35}, \bibinfo{number}{4} (\bibinfo{year}{2016}),
  \bibinfo{pages}{1--13}.
\newblock


\bibitem[Gao et~al\mbox{.}(2022)]%
        {gaoMPSNeRFGeneralizable3D2022}
\bibfield{author}{\bibinfo{person}{Xiangjun Gao}, \bibinfo{person}{Jiaolong
  Yang}, \bibinfo{person}{Jongyoo Kim}, \bibinfo{person}{Sida Peng},
  \bibinfo{person}{Zicheng Liu}, {and} \bibinfo{person}{Xin Tong}.}
  \bibinfo{year}{2022}\natexlab{}.
\newblock \showarticletitle{{{MPS-NeRF}}: {{Generalizable 3D Human Rendering
  From Multiview Images}}}.
\newblock \bibinfo{journal}{\emph{TPAMI}}  \bibinfo{volume}{PP}
  (\bibinfo{date}{Sept.} \bibinfo{year}{2022}).
\newblock
\showISSN{1939-3539}


\bibitem[Guo et~al\mbox{.}(2019)]%
        {guoRelightablesVolumetricPerformance2019}
\bibfield{author}{\bibinfo{person}{Kaiwen Guo}, \bibinfo{person}{Peter
  Lincoln}, \bibinfo{person}{Philip Davidson}, \bibinfo{person}{Jay Busch},
  \bibinfo{person}{Xueming Yu}, \bibinfo{person}{Matt Whalen},
  \bibinfo{person}{Geoff Harvey}, \bibinfo{person}{Sergio {Orts-Escolano}},
  \bibinfo{person}{Rohit Pandey}, {and} \bibinfo{person}{Jason Dourgarian}.}
  \bibinfo{year}{2019}\natexlab{}.
\newblock \showarticletitle{The Relightables: {{Volumetric}} Performance
  Capture of Humans with Realistic Relighting}.
\newblock \bibinfo{journal}{\emph{ACM Transactions on Graphics (ToG)}}
  \bibinfo{volume}{38}, \bibinfo{number}{6} (\bibinfo{year}{2019}),
  \bibinfo{pages}{1--19}.
\newblock


\bibitem[He et~al\mbox{.}(2021)]%
        {heARCHAnimationreadyClothed2021}
\bibfield{author}{\bibinfo{person}{Tong He}, \bibinfo{person}{Yuanlu Xu},
  \bibinfo{person}{Shunsuke Saito}, \bibinfo{person}{Stefano Soatto}, {and}
  \bibinfo{person}{Tony Tung}.} \bibinfo{year}{2021}\natexlab{}.
\newblock \showarticletitle{{{ARCH}}++: {{Animation-ready}} Clothed Human
  Reconstruction Revisited}. In \bibinfo{booktitle}{\emph{ICCV}}.
  \bibinfo{pages}{11046--11056}.
\newblock


\bibitem[Huang et~al\mbox{.}(2020)]%
        {huangArchAnimatableReconstruction2020}
\bibfield{author}{\bibinfo{person}{Zeng Huang}, \bibinfo{person}{Yuanlu Xu},
  \bibinfo{person}{Christoph Lassner}, \bibinfo{person}{Hao Li}, {and}
  \bibinfo{person}{Tony Tung}.} \bibinfo{year}{2020}\natexlab{}.
\newblock \showarticletitle{Arch: {{Animatable}} Reconstruction of Clothed
  Humans}. In \bibinfo{booktitle}{\emph{CVPR}}. \bibinfo{pages}{3093--3102}.
\newblock


\bibitem[Ionescu et~al\mbox{.}(2013)]%
        {ionescuHuman36mLarge2013}
\bibfield{author}{\bibinfo{person}{Catalin Ionescu}, \bibinfo{person}{Dragos
  Papava}, \bibinfo{person}{Vlad Olaru}, {and} \bibinfo{person}{Cristian
  Sminchisescu}.} \bibinfo{year}{2013}\natexlab{}.
\newblock \showarticletitle{Human3. 6m: {{Large}} Scale Datasets and Predictive
  Methods for 3d Human Sensing in Natural Environments}.
\newblock \bibinfo{journal}{\emph{TPAMI}} \bibinfo{volume}{36},
  \bibinfo{number}{7} (\bibinfo{year}{2013}), \bibinfo{pages}{1325--1339}.
\newblock


\bibitem[Kajiya and Von~Herzen(1984)]%
        {kajiyaRayTracingVolume1984}
\bibfield{author}{\bibinfo{person}{James~T. Kajiya} {and}
  \bibinfo{person}{Brian~P. Von~Herzen}.} \bibinfo{year}{1984}\natexlab{}.
\newblock \showarticletitle{Ray Tracing Volume Densities}.
\newblock \bibinfo{journal}{\emph{ACM SIGGRAPH Computer Graphics}}
  \bibinfo{volume}{18}, \bibinfo{number}{3} (\bibinfo{year}{1984}),
  \bibinfo{pages}{165--174}.
\newblock


\bibitem[Kavan et~al\mbox{.}(2007)]%
        {kavanSkinningDualQuaternions2007}
\bibfield{author}{\bibinfo{person}{Ladislav Kavan}, \bibinfo{person}{Steven
  Collins}, \bibinfo{person}{Ji{\v r}{\'i} {\v Z}{\'a}ra}, {and}
  \bibinfo{person}{Carol O'Sullivan}.} \bibinfo{year}{2007}\natexlab{}.
\newblock \showarticletitle{Skinning with Dual Quaternions}. In
  \bibinfo{booktitle}{\emph{Proceedings of the 2007 Symposium on {{Interactive
  3D}} Graphics and Games}}. \bibinfo{pages}{39--46}.
\newblock


\bibitem[Kingma and Ba(2015)]%
        {kingmaAdamMethodStochastic2015}
\bibfield{author}{\bibinfo{person}{Diederik~P. Kingma} {and}
  \bibinfo{person}{Jimmy Ba}.} \bibinfo{year}{2015}\natexlab{}.
\newblock \showarticletitle{Adam: {{A Method}} for {{Stochastic
  Optimization}}}. In \bibinfo{booktitle}{\emph{{{ICLR}} ({{Poster}})}}.
\newblock


\bibitem[Kwon et~al\mbox{.}(2021)]%
        {kwonNeuralHumanPerformer2021}
\bibfield{author}{\bibinfo{person}{Youngjoong Kwon}, \bibinfo{person}{Dahun
  Kim}, \bibinfo{person}{Duygu Ceylan}, {and} \bibinfo{person}{Henry Fuchs}.}
  \bibinfo{year}{2021}\natexlab{}.
\newblock \showarticletitle{Neural {{Human Performer}}: {{Learning
  Generalizable Radiance Fields}} for {{Human Performance Rendering}}}. In
  \bibinfo{booktitle}{\emph{Advances in {{Neural Information Processing
  Systems}}}}, Vol.~\bibinfo{volume}{34}. \bibinfo{publisher}{{Curran
  Associates, Inc.}}, \bibinfo{pages}{24741--24752}.
\newblock


\bibitem[Lewis et~al\mbox{.}(2000)]%
        {lewisPoseSpaceDeformation2000a}
\bibfield{author}{\bibinfo{person}{John~P. Lewis}, \bibinfo{person}{Matt
  Cordner}, {and} \bibinfo{person}{Nickson Fong}.}
  \bibinfo{year}{2000}\natexlab{}.
\newblock \showarticletitle{Pose Space Deformation: A Unified Approach to Shape
  Interpolation and Skeleton-Driven Deformation}. In
  \bibinfo{booktitle}{\emph{Proceedings of the 27th Annual Conference on
  {{Computer}} Graphics and Interactive Techniques}}.
  \bibinfo{pages}{165--172}.
\newblock


\bibitem[Li et~al\mbox{.}(2021)]%
        {liNeuralSceneFlow2021}
\bibfield{author}{\bibinfo{person}{Zhengqi Li}, \bibinfo{person}{Simon
  Niklaus}, \bibinfo{person}{Noah Snavely}, {and} \bibinfo{person}{Oliver
  Wang}.} \bibinfo{year}{2021}\natexlab{}.
\newblock \showarticletitle{Neural Scene Flow Fields for Space-Time View
  Synthesis of Dynamic Scenes}. In \bibinfo{booktitle}{\emph{CVPR}}.
  \bibinfo{pages}{6498--6508}.
\newblock


\bibitem[Liu et~al\mbox{.}(2021)]%
        {liuNeuralActorNeural2021}
\bibfield{author}{\bibinfo{person}{Lingjie Liu}, \bibinfo{person}{Marc
  Habermann}, \bibinfo{person}{Viktor Rudnev}, \bibinfo{person}{Kripasindhu
  Sarkar}, \bibinfo{person}{Jiatao Gu}, {and} \bibinfo{person}{Christian
  Theobalt}.} \bibinfo{year}{2021}\natexlab{}.
\newblock \showarticletitle{Neural Actor: {{Neural}} Free-View Synthesis of
  Human Actors with Pose Control}.
\newblock \bibinfo{journal}{\emph{ACM Transactions on Graphics (TOG)}}
  \bibinfo{volume}{40}, \bibinfo{number}{6} (\bibinfo{year}{2021}),
  \bibinfo{pages}{1--16}.
\newblock


\bibitem[Lombardi et~al\mbox{.}(2019)]%
        {lombardiNeuralVolumesLearning2019}
\bibfield{author}{\bibinfo{person}{Stephen Lombardi}, \bibinfo{person}{Tomas
  Simon}, \bibinfo{person}{Jason Saragih}, \bibinfo{person}{Gabriel Schwartz},
  \bibinfo{person}{Andreas Lehrmann}, {and} \bibinfo{person}{Yaser Sheikh}.}
  \bibinfo{year}{2019}\natexlab{}.
\newblock \showarticletitle{Neural {{Volumes}}: {{Learning Dynamic Renderable
  Volumes}} from {{Images}}}.
\newblock \bibinfo{journal}{\emph{ACM Transactions on Graphics (TOG)}}
  \bibinfo{volume}{38}, \bibinfo{number}{4} (\bibinfo{date}{Aug.}
  \bibinfo{year}{2019}), \bibinfo{pages}{1--14}.
\newblock
\showISSN{0730-0301, 1557-7368}
\showeprint[arxiv]{1906.07751}~[cs]


\bibitem[Loper et~al\mbox{.}(2015)]%
        {loperSMPLSkinnedMultiperson2015}
\bibfield{author}{\bibinfo{person}{Matthew Loper}, \bibinfo{person}{Naureen
  Mahmood}, \bibinfo{person}{Javier Romero}, \bibinfo{person}{Gerard
  {Pons-Moll}}, {and} \bibinfo{person}{Michael~J. Black}.}
  \bibinfo{year}{2015}\natexlab{}.
\newblock \showarticletitle{{{SMPL}}: {{A}} Skinned Multi-Person Linear Model}.
\newblock \bibinfo{journal}{\emph{ACM Transactions on Graphics (TOG)}}
  \bibinfo{volume}{34}, \bibinfo{number}{6} (\bibinfo{year}{2015}),
  \bibinfo{pages}{1--16}.
\newblock


\bibitem[Max(1995)]%
        {maxOpticalModelsDirect1995}
\bibfield{author}{\bibinfo{person}{Nelson Max}.}
  \bibinfo{year}{1995}\natexlab{}.
\newblock \showarticletitle{Optical Models for Direct Volume Rendering}.
\newblock \bibinfo{journal}{\emph{IEEE Transactions on Visualization and
  Computer Graphics}} \bibinfo{volume}{1}, \bibinfo{number}{2}
  (\bibinfo{year}{1995}), \bibinfo{pages}{99--108}.
\newblock


\bibitem[Mescheder et~al\mbox{.}(2019)]%
        {meschederOccupancyNetworksLearning2019}
\bibfield{author}{\bibinfo{person}{Lars Mescheder}, \bibinfo{person}{Michael
  Oechsle}, \bibinfo{person}{Michael Niemeyer}, \bibinfo{person}{Sebastian
  Nowozin}, {and} \bibinfo{person}{Andreas Geiger}.}
  \bibinfo{year}{2019}\natexlab{}.
\newblock \showarticletitle{Occupancy Networks: {{Learning}} 3d Reconstruction
  in Function Space}. In \bibinfo{booktitle}{\emph{CVPR}}.
  \bibinfo{pages}{4460--4470}.
\newblock


\bibitem[Mihajlovic et~al\mbox{.}(2022)]%
        {mihajlovicKeypointNeRFGeneralizingImagebased2022}
\bibfield{author}{\bibinfo{person}{Marko Mihajlovic}, \bibinfo{person}{Aayush
  Bansal}, \bibinfo{person}{Michael Zollhoefer}, \bibinfo{person}{Siyu Tang},
  {and} \bibinfo{person}{Shunsuke Saito}.} \bibinfo{year}{2022}\natexlab{}.
\newblock \showarticletitle{{KeypointNeRF}: Generalizing Image-based Volumetric
  Avatars using Relative Spatial Encoding of Keypoints}. In
  \bibinfo{booktitle}{\emph{ECCV}}.
\newblock


\bibitem[Mildenhall et~al\mbox{.}(2021)]%
        {mildenhallNerfRepresentingScenes2021}
\bibfield{author}{\bibinfo{person}{Ben Mildenhall}, \bibinfo{person}{Pratul~P.
  Srinivasan}, \bibinfo{person}{Matthew Tancik}, \bibinfo{person}{Jonathan~T.
  Barron}, \bibinfo{person}{Ravi Ramamoorthi}, {and} \bibinfo{person}{Ren Ng}.}
  \bibinfo{year}{2021}\natexlab{}.
\newblock \showarticletitle{Nerf: {{Representing}} Scenes as Neural Radiance
  Fields for View Synthesis}.
\newblock \bibinfo{journal}{\emph{Commun. ACM}} \bibinfo{volume}{65},
  \bibinfo{number}{1} (\bibinfo{year}{2021}), \bibinfo{pages}{99--106}.
\newblock


\bibitem[M{\"u}ller et~al\mbox{.}(2022)]%
        {mullerInstantNeuralGraphics2022}
\bibfield{author}{\bibinfo{person}{Thomas M{\"u}ller}, \bibinfo{person}{Alex
  Evans}, \bibinfo{person}{Christoph Schied}, {and} \bibinfo{person}{Alexander
  Keller}.} \bibinfo{year}{2022}\natexlab{}.
\newblock \showarticletitle{Instant Neural Graphics Primitives with a
  Multiresolution Hash Encoding}.
\newblock \bibinfo{journal}{\emph{ACM Transactions on Graphics (TOG)}}
  \bibinfo{volume}{41}, \bibinfo{number}{4} (\bibinfo{date}{July}
  \bibinfo{year}{2022}), \bibinfo{pages}{102:1--102:15}.
\newblock
\showISSN{0730-0301}


\bibitem[Noguchi et~al\mbox{.}(2021)]%
        {noguchiNeuralArticulatedRadiance2021}
\bibfield{author}{\bibinfo{person}{Atsuhiro Noguchi}, \bibinfo{person}{Xiao
  Sun}, \bibinfo{person}{Stephen Lin}, {and} \bibinfo{person}{Tatsuya Harada}.}
  \bibinfo{year}{2021}\natexlab{}.
\newblock \showarticletitle{Neural Articulated Radiance Field}. In
  \bibinfo{booktitle}{\emph{ICCV}}. \bibinfo{pages}{5762--5772}.
\newblock


\bibitem[Park et~al\mbox{.}(2019)]%
        {parkDeepsdfLearningContinuous2019}
\bibfield{author}{\bibinfo{person}{Jeong~Joon Park}, \bibinfo{person}{Peter
  Florence}, \bibinfo{person}{Julian Straub}, \bibinfo{person}{Richard
  Newcombe}, {and} \bibinfo{person}{Steven Lovegrove}.}
  \bibinfo{year}{2019}\natexlab{}.
\newblock \showarticletitle{Deepsdf: {{Learning}} Continuous Signed Distance
  Functions for Shape Representation}. In \bibinfo{booktitle}{\emph{CVPR}}.
  \bibinfo{pages}{165--174}.
\newblock


\bibitem[Park et~al\mbox{.}(2021)]%
        {parkNerfiesDeformableNeural2021}
\bibfield{author}{\bibinfo{person}{Keunhong Park}, \bibinfo{person}{Utkarsh
  Sinha}, \bibinfo{person}{Jonathan~T. Barron}, \bibinfo{person}{Sofien
  Bouaziz}, \bibinfo{person}{Dan~B. Goldman}, \bibinfo{person}{Steven~M.
  Seitz}, {and} \bibinfo{person}{Ricardo {Martin-Brualla}}.}
  \bibinfo{year}{2021}\natexlab{}.
\newblock \showarticletitle{Nerfies: {{Deformable}} Neural Radiance Fields}. In
  \bibinfo{booktitle}{\emph{ICCV}}. \bibinfo{pages}{5865--5874}.
\newblock


\bibitem[Paszke et~al\mbox{.}(2017)]%
        {paszke2017automatic}
\bibfield{author}{\bibinfo{person}{Adam Paszke}, \bibinfo{person}{Sam Gross},
  \bibinfo{person}{Soumith Chintala}, \bibinfo{person}{Gregory Chanan},
  \bibinfo{person}{Edward Yang}, \bibinfo{person}{Zachary DeVito},
  \bibinfo{person}{Zeming Lin}, \bibinfo{person}{Alban Desmaison},
  \bibinfo{person}{Luca Antiga}, {and} \bibinfo{person}{Adam Lerer}.}
  \bibinfo{year}{2017}\natexlab{}.
\newblock \showarticletitle{Automatic differentiation in PyTorch}.
\newblock  (\bibinfo{year}{2017}).
\newblock


\bibitem[Peng et~al\mbox{.}(2021a)]%
        {pengAnimatableNeuralRadiance2021a}
\bibfield{author}{\bibinfo{person}{Sida Peng}, \bibinfo{person}{Junting Dong},
  \bibinfo{person}{Qianqian Wang}, \bibinfo{person}{Shangzhan Zhang},
  \bibinfo{person}{Qing Shuai}, \bibinfo{person}{Xiaowei Zhou}, {and}
  \bibinfo{person}{Hujun Bao}.} \bibinfo{year}{2021}\natexlab{a}.
\newblock \showarticletitle{Animatable Neural Radiance Fields for Modeling
  Dynamic Human Bodies}. In \bibinfo{booktitle}{\emph{ICCV}}.
  \bibinfo{pages}{14314--14323}.
\newblock


\bibitem[Peng et~al\mbox{.}(2021b)]%
        {pengNeuralBodyImplicit2021}
\bibfield{author}{\bibinfo{person}{Sida Peng}, \bibinfo{person}{Yuanqing
  Zhang}, \bibinfo{person}{Yinghao Xu}, \bibinfo{person}{Qianqian Wang},
  \bibinfo{person}{Qing Shuai}, \bibinfo{person}{Hujun Bao}, {and}
  \bibinfo{person}{Xiaowei Zhou}.} \bibinfo{year}{2021}\natexlab{b}.
\newblock \showarticletitle{Neural Body: {{Implicit}} Neural Representations
  with Structured Latent Codes for Novel View Synthesis of Dynamic Humans}. In
  \bibinfo{booktitle}{\emph{CVPR}}. \bibinfo{pages}{9054--9063}.
\newblock


\bibitem[Pumarola et~al\mbox{.}(2021)]%
        {pumarolaDnerfNeuralRadiance2021}
\bibfield{author}{\bibinfo{person}{Albert Pumarola}, \bibinfo{person}{Enric
  Corona}, \bibinfo{person}{Gerard {Pons-Moll}}, {and}
  \bibinfo{person}{Francesc {Moreno-Noguer}}.} \bibinfo{year}{2021}\natexlab{}.
\newblock \showarticletitle{D-Nerf: {{Neural}} Radiance Fields for Dynamic
  Scenes}. In \bibinfo{booktitle}{\emph{CVPR}}. \bibinfo{pages}{10318--10327}.
\newblock


\bibitem[Raj et~al\mbox{.}(2021a)]%
        {rajAnrArticulatedNeural2021}
\bibfield{author}{\bibinfo{person}{Amit Raj}, \bibinfo{person}{Julian Tanke},
  \bibinfo{person}{James Hays}, \bibinfo{person}{Minh Vo},
  \bibinfo{person}{Carsten Stoll}, {and} \bibinfo{person}{Christoph Lassner}.}
  \bibinfo{year}{2021}\natexlab{a}.
\newblock \showarticletitle{Anr: {{Articulated}} Neural Rendering for Virtual
  Avatars}. In \bibinfo{booktitle}{\emph{CVPR}}. \bibinfo{pages}{3722--3731}.
\newblock


\bibitem[Raj et~al\mbox{.}(2021b)]%
        {rajPvaPixelalignedVolumetric2021}
\bibfield{author}{\bibinfo{person}{Amit Raj}, \bibinfo{person}{Michael
  Zollhoefer}, \bibinfo{person}{Tomas Simon}, \bibinfo{person}{Jason Saragih},
  \bibinfo{person}{Shunsuke Saito}, \bibinfo{person}{James Hays}, {and}
  \bibinfo{person}{Stephen Lombardi}.} \bibinfo{year}{2021}\natexlab{b}.
\newblock \showarticletitle{Pva: {{Pixel-aligned}} Volumetric Avatars}.
\newblock \bibinfo{journal}{\emph{arXiv preprint arXiv:2101.02697}}
  (\bibinfo{year}{2021}).
\newblock
\showeprint[arxiv]{2101.02697}


\bibitem[Saito et~al\mbox{.}(2019)]%
        {saitoPifuPixelalignedImplicit2019}
\bibfield{author}{\bibinfo{person}{Shunsuke Saito}, \bibinfo{person}{Zeng
  Huang}, \bibinfo{person}{Ryota Natsume}, \bibinfo{person}{Shigeo Morishima},
  \bibinfo{person}{Angjoo Kanazawa}, {and} \bibinfo{person}{Hao Li}.}
  \bibinfo{year}{2019}\natexlab{}.
\newblock \showarticletitle{Pifu: {{Pixel-aligned}} Implicit Function for
  High-Resolution Clothed Human Digitization}. In
  \bibinfo{booktitle}{\emph{ICCV}}. \bibinfo{pages}{2304--2314}.
\newblock


\bibitem[Saito et~al\mbox{.}(2020)]%
        {saitoPifuhdMultilevelPixelaligned2020}
\bibfield{author}{\bibinfo{person}{Shunsuke Saito}, \bibinfo{person}{Tomas
  Simon}, \bibinfo{person}{Jason Saragih}, {and} \bibinfo{person}{Hanbyul
  Joo}.} \bibinfo{year}{2020}\natexlab{}.
\newblock \showarticletitle{Pifuhd: {{Multi-level}} Pixel-Aligned Implicit
  Function for High-Resolution 3d Human Digitization}. In
  \bibinfo{booktitle}{\emph{CVPR}}. \bibinfo{pages}{84--93}.
\newblock


\bibitem[Simonyan and Zisserman(2015)]%
        {simonyanVeryDeepConvolutional2015}
\bibfield{author}{\bibinfo{person}{Karen Simonyan} {and}
  \bibinfo{person}{Andrew Zisserman}.} \bibinfo{year}{2015}\natexlab{}.
\newblock \showarticletitle{Very {{Deep Convolutional Networks}} for
  {{Large-Scale Image Recognition}}}. In \bibinfo{booktitle}{\emph{ICLR}}.
\newblock


\bibitem[Sitzmann et~al\mbox{.}(2019)]%
        {sitzmannSceneRepresentationNetworks2019}
\bibfield{author}{\bibinfo{person}{Vincent Sitzmann}, \bibinfo{person}{Michael
  Zollh{\"o}fer}, {and} \bibinfo{person}{Gordon Wetzstein}.}
  \bibinfo{year}{2019}\natexlab{}.
\newblock \showarticletitle{Scene Representation Networks: {{Continuous}}
  3d-Structure-Aware Neural Scene Representations}.
\newblock \bibinfo{journal}{\emph{Advances in Neural Information Processing
  Systems}}  \bibinfo{volume}{32} (\bibinfo{year}{2019}).
\newblock


\bibitem[Su et~al\mbox{.}(2021)]%
        {suAnerfArticulatedNeural2021}
\bibfield{author}{\bibinfo{person}{Shih-Yang Su}, \bibinfo{person}{Frank Yu},
  \bibinfo{person}{Michael Zollh{\"o}fer}, {and} \bibinfo{person}{Helge
  Rhodin}.} \bibinfo{year}{2021}\natexlab{}.
\newblock \showarticletitle{A-Nerf: {{Articulated}} Neural Radiance Fields for
  Learning Human Shape, Appearance, and Pose}.
\newblock \bibinfo{journal}{\emph{Advances in Neural Information Processing
  Systems}}  \bibinfo{volume}{34} (\bibinfo{year}{2021}),
  \bibinfo{pages}{12278--12291}.
\newblock


\bibitem[Tancik et~al\mbox{.}(2020)]%
        {tancikFourierFeaturesLet2020}
\bibfield{author}{\bibinfo{person}{Matthew Tancik}, \bibinfo{person}{Pratul
  Srinivasan}, \bibinfo{person}{Ben Mildenhall}, \bibinfo{person}{Sara
  {Fridovich-Keil}}, \bibinfo{person}{Nithin Raghavan},
  \bibinfo{person}{Utkarsh Singhal}, \bibinfo{person}{Ravi Ramamoorthi},
  \bibinfo{person}{Jonathan Barron}, {and} \bibinfo{person}{Ren Ng}.}
  \bibinfo{year}{2020}\natexlab{}.
\newblock \showarticletitle{Fourier Features Let Networks Learn High Frequency
  Functions in Low Dimensional Domains}.
\newblock \bibinfo{journal}{\emph{Advances in Neural Information Processing
  Systems}}  \bibinfo{volume}{33} (\bibinfo{year}{2020}),
  \bibinfo{pages}{7537--7547}.
\newblock


\bibitem[Tretschk et~al\mbox{.}(2021)]%
        {tretschkNonrigidNeuralRadiance2021}
\bibfield{author}{\bibinfo{person}{Edgar Tretschk}, \bibinfo{person}{Ayush
  Tewari}, \bibinfo{person}{Vladislav Golyanik}, \bibinfo{person}{Michael
  Zollh{\"o}fer}, \bibinfo{person}{Christoph Lassner}, {and}
  \bibinfo{person}{Christian Theobalt}.} \bibinfo{year}{2021}\natexlab{}.
\newblock \showarticletitle{Non-Rigid Neural Radiance Fields:
  {{Reconstruction}} and Novel View Synthesis of a Dynamic Scene from Monocular
  Video}. In \bibinfo{booktitle}{\emph{ICCV}}. \bibinfo{pages}{12959--12970}.
\newblock


\bibitem[Trevithick and Yang(2021)]%
        {trevithickGrfLearningGeneral2021}
\bibfield{author}{\bibinfo{person}{Alex Trevithick} {and} \bibinfo{person}{Bo
  Yang}.} \bibinfo{year}{2021}\natexlab{}.
\newblock \showarticletitle{Grf: {{Learning}} a General Radiance Field for 3d
  Representation and Rendering}. In \bibinfo{booktitle}{\emph{ICCV}}.
  \bibinfo{pages}{15182--15192}.
\newblock


\bibitem[Wang et~al\mbox{.}(2021)]%
        {wangIbrnetLearningMultiview2021}
\bibfield{author}{\bibinfo{person}{Qianqian Wang}, \bibinfo{person}{Zhicheng
  Wang}, \bibinfo{person}{Kyle Genova}, \bibinfo{person}{Pratul~P. Srinivasan},
  \bibinfo{person}{Howard Zhou}, \bibinfo{person}{Jonathan~T. Barron},
  \bibinfo{person}{Ricardo {Martin-Brualla}}, \bibinfo{person}{Noah Snavely},
  {and} \bibinfo{person}{Thomas Funkhouser}.} \bibinfo{year}{2021}\natexlab{}.
\newblock \showarticletitle{IBRnet: {{Learning}} Multi-View Image-Based
  Rendering}. In \bibinfo{booktitle}{\emph{CVPR}}. \bibinfo{pages}{4690--4699}.
\newblock


\bibitem[Wang et~al\mbox{.}(2022)]%
        {wangNeuralNovelActor2022}
\bibfield{author}{\bibinfo{person}{Yiming Wang}, \bibinfo{person}{Qingzhe Gao},
  \bibinfo{person}{Libin Liu}, \bibinfo{person}{Lingjie Liu},
  \bibinfo{person}{Christian Theobalt}, {and} \bibinfo{person}{Baoquan Chen}.}
  \bibinfo{year}{2022}\natexlab{}.
\newblock \bibinfo{title}{Neural {{Novel Actor}}: {{Learning}} a {{Generalized
  Animatable Neural Representation}} for {{Human Actors}}}.
\newblock
\newblock
\showeprint[arxiv]{2208.11905}~[cs]


\bibitem[Weng et~al\mbox{.}(2022)]%
        {wengHumannerfFreeviewpointRendering2022}
\bibfield{author}{\bibinfo{person}{Chung-Yi Weng}, \bibinfo{person}{Brian
  Curless}, \bibinfo{person}{Pratul~P. Srinivasan},
  \bibinfo{person}{Jonathan~T. Barron}, {and} \bibinfo{person}{Ira
  {Kemelmacher-Shlizerman}}.} \bibinfo{year}{2022}\natexlab{}.
\newblock \showarticletitle{Humannerf: {{Free-viewpoint}} Rendering of Moving
  People from Monocular Video}. In \bibinfo{booktitle}{\emph{CVPR}}.
  \bibinfo{pages}{16210--16220}.
\newblock


\bibitem[Wu et~al\mbox{.}(2020)]%
        {wuMultiviewNeuralHuman2020}
\bibfield{author}{\bibinfo{person}{Minye Wu}, \bibinfo{person}{Yuehao Wang},
  \bibinfo{person}{Qiang Hu}, {and} \bibinfo{person}{Jingyi Yu}.}
  \bibinfo{year}{2020}\natexlab{}.
\newblock \showarticletitle{Multi-View Neural Human Rendering}. In
  \bibinfo{booktitle}{\emph{CVPR}}. \bibinfo{pages}{1682--1691}.
\newblock


\bibitem[Yang et~al\mbox{.}(2021)]%
        {yangS3NeuralShape2021}
\bibfield{author}{\bibinfo{person}{Ze Yang}, \bibinfo{person}{Shenlong Wang},
  \bibinfo{person}{Sivabalan Manivasagam}, \bibinfo{person}{Zeng Huang},
  \bibinfo{person}{Wei-Chiu Ma}, \bibinfo{person}{Xinchen Yan},
  \bibinfo{person}{Ersin Yumer}, {and} \bibinfo{person}{Raquel Urtasun}.}
  \bibinfo{year}{2021}\natexlab{}.
\newblock \showarticletitle{S3: {{Neural}} Shape, Skeleton, and Skinning Fields
  for 3d Human Modeling}. In \bibinfo{booktitle}{\emph{CVPR}}.
  \bibinfo{pages}{13284--13293}.
\newblock


\bibitem[Yariv et~al\mbox{.}(2021)]%
        {yarivVolumeRenderingNeural2021}
\bibfield{author}{\bibinfo{person}{Lior Yariv}, \bibinfo{person}{Jiatao Gu},
  \bibinfo{person}{Yoni Kasten}, {and} \bibinfo{person}{Yaron Lipman}.}
  \bibinfo{year}{2021}\natexlab{}.
\newblock \showarticletitle{Volume Rendering of Neural Implicit Surfaces}.
\newblock \bibinfo{journal}{\emph{Advances in Neural Information Processing
  Systems}}  \bibinfo{volume}{34} (\bibinfo{year}{2021}),
  \bibinfo{pages}{4805--4815}.
\newblock


\bibitem[Yu et~al\mbox{.}(2021)]%
        {yuPixelnerfNeuralRadiance2021}
\bibfield{author}{\bibinfo{person}{Alex Yu}, \bibinfo{person}{Vickie Ye},
  \bibinfo{person}{Matthew Tancik}, {and} \bibinfo{person}{Angjoo Kanazawa}.}
  \bibinfo{year}{2021}\natexlab{}.
\newblock \showarticletitle{Pixelnerf: {{Neural}} Radiance Fields from One or
  Few Images}. In \bibinfo{booktitle}{\emph{CVPR}}.
  \bibinfo{pages}{4578--4587}.
\newblock


\bibitem[Zhang and Tao(2020)]%
        {zhang2020empowering}
\bibfield{author}{\bibinfo{person}{Jing Zhang} {and} \bibinfo{person}{Dacheng
  Tao}.} \bibinfo{year}{2020}\natexlab{}.
\newblock \showarticletitle{Empowering things with intelligence: a survey of
  the progress, challenges, and opportunities in artificial intelligence of
  things}.
\newblock \bibinfo{journal}{\emph{IEEE Internet of Things Journal}}
  \bibinfo{volume}{8}, \bibinfo{number}{10} (\bibinfo{year}{2020}),
  \bibinfo{pages}{7789--7817}.
\newblock


\bibitem[Zhao et~al\mbox{.}(2022)]%
        {zhaoHumanNeRFEfficientlyGenerated2022}
\bibfield{author}{\bibinfo{person}{Fuqiang Zhao}, \bibinfo{person}{Wei Yang},
  \bibinfo{person}{Jiakai Zhang}, \bibinfo{person}{Pei Lin},
  \bibinfo{person}{Yingliang Zhang}, \bibinfo{person}{Jingyi Yu}, {and}
  \bibinfo{person}{Lan Xu}.} \bibinfo{year}{2022}\natexlab{}.
\newblock \showarticletitle{{{HumanNeRF}}: {{Efficiently Generated Human
  Radiance Field}} from {{Sparse Inputs}}}. In
  \bibinfo{booktitle}{\emph{CVPR}}. \bibinfo{pages}{7743--7753}.
\newblock


\bibitem[Zhi et~al\mbox{.}(2022)]%
        {dual-nerf}
\bibfield{author}{\bibinfo{person}{Y. Zhi}, \bibinfo{person}{S. Qian},
  \bibinfo{person}{X. Yan}, {and} \bibinfo{person}{S. Gao}.}
  \bibinfo{year}{2022}\natexlab{}.
\newblock \showarticletitle{Dual-Space NeRF: Learning Animatable Avatars and
  Scene Lighting in Separate Spaces}. In \bibinfo{booktitle}{\emph{2022
  International Conference on 3D Vision (3DV)}}. \bibinfo{publisher}{IEEE
  Computer Society}, \bibinfo{address}{Los Alamitos, CA, USA},
  \bibinfo{pages}{1--10}.
\newblock


\end{thebibliography}

\end{document}